\documentclass{article} % For LaTeX2e
\usepackage{iclr2022_conference_noheader,times}

\usepackage{amsmath,amsfonts,bm}

\def\eqref#1{equation~\ref{#1}}
\def\1{\bm{1}}

\DeclareMathAlphabet{\mathsfit}{\encodingdefault}{\sfdefault}{m}{sl}
\SetMathAlphabet{\mathsfit}{bold}{\encodingdefault}{\sfdefault}{bx}{n}

\usepackage{hyperref}
\usepackage{url}

\usepackage{caption}
\usepackage{subcaption}

\usepackage[pdftex]{graphicx}

\title{Pre-training Co-evolutionary Protein Representation via A Pairwise Masked Language Model}

\author{Liang He  \\
	Microsoft Research Asia\\
	Beijing, China \\
	\textit{lihe@microsoft.com} \\
	\And
	Shizhuo Zhang\thanks{This work was conducted at Microsoft Research Asia}  \\
	Nanyang Technological University \\
	Jurong West, Singapore \\
	\textit{M180175@e.ntu.edu.sg} \\
	\And
	Lijun Wu  \\
	Microsoft Research Asia\\
	Beijing, China \\
	\textit{lijuwu@microsoft.com} \\
	\And
	Huanhuan Xia, Fusong Ju  \\
	Microsoft Research Asia\\
	Beijing, China \\
	\textit{\{lexi,fusongju\}@microsoft.com} \\
	\And
	He Zhang$^*$ \\
	Xi'an Jiaotong University \\
	Shaanxi, China \\
	\textit{mao736488798@stu.xjtu.edu.cn} 
	\And
	Siyuan Liu$^*$  \\
	Sun Yat-sen University \\
	Guangdong, China \\
	\textit{liusy8@mail2.sysu.edu.cn} \\
	\And
	Yingce Xia, Jianwei Zhu, Pan Deng, Bin Shao, Tao Qin, Tie-Yan Liu \\
	Microsoft Research Asia\\
	Beijing, China \\
	\textit{\{yinxia,jianwzhu,paden,binshao,taoqin,tyliu\}@microsoft.com} \\
}

\newcommand{\TableSize}{\small}

\iclrfinalcopy % Uncomment for camera-ready version, but NOT for submission.

\begin{document}

\maketitle

\begin{abstract}
%Life is ruled by biological sequences and molecules, especially protein sequences, following the de facto language of life.
Understanding protein sequences is vital and urgent for biology, healthcare, and medicine.
Labeling approaches are expensive yet time-consuming, 
while the amount of unlabeled data is increasing quite faster than that of the labeled data
due to low-cost, high-throughput sequencing methods.
In order to extract knowledge from these unlabeled data, 
representation learning is of significant value for protein-related tasks and 
has great potential for helping us learn more about protein functions and structures.
%		Given that many pre-training methods for protein sequences have been proposed, 
%		them are considered straightforward adaptations of the NLP pre-training methods to protein sequences, 
%		typically by applying the well-acknowledged natural language models to pre-training.
%		According to Anfinsen's dogma for protein, the structure is determined only by the sequence, 
%		and its function is realized by the structure.
The key problem in the protein sequence representation learning is to capture 
the co-evolutionary information reflected by the inter-residue co-variation in the sequences.
%		The goal is usually achieved by leveraging multiple sequence alignment, 
%		which essentially grouping the similar sequences to generate the representation.
Instead of leveraging multiple sequence alignment as is usually done, % or applying NLP originated masked language model,
we propose a novel method to capture this information directly by pre-training
via a dedicated language model, i.e., \textit{Pairwise Masked Language Model (PMLM)}.
In a conventional masked language model, the masked tokens (i.e. amino acid residues) are modeled by conditioning on the unmasked tokens only,
but processed independently to each other. However, our proposed PMLM takes the dependency among masked tokens into consideration, 
i.e., the probability of a token pair is not equal to the product of the probability of the two tokens.
By applying this model, the pre-trained encoder is able to generate a better representation for protein sequences.
%that likely better captures the co-evolutionary information .
Our result shows that the proposed method can effectively capture the inter-residue correlations and
improves the performance of contact prediction by up to $9\%$ compared to the MLM baseline under the same setting.
The proposed model also significantly outperforms the MSA baseline by more than $7\%$	 on the TAPE contact prediction benchmark 
when pre-trained on a subset of the sequence database which the MSA is generated from, % a more diverse sequence dataset,
revealing the potential of the sequence pre-training method to surpass MSA based methods in general.
%The result shows that the performance is not hurt by the design
%when evaluated on two other downstream tasks.
\end{abstract}

\section{Introduction}

Life is ruled by biological sequences and molecules, i.e. DNA, RNA, and protein sequences, 
following the de facto ‘natural’ language of biology. 
For protein, the structure is determined by the sequence,
and its function is realized by the structure, according to Anfinsen's dogma. 
However, the structure and function label is never easy to obtain (time-consuming and expensive) 
nor always effective due to the dark regions and structure dynamics.
%(dark proteome consisted of Intrinsically Disordered Proteins are widely existed, 
%e.g., 44-54\% of the proteome in eukaryotes and viruses, 
%while a lot of protein sequences have few homologous sequences). 
%Although sequence analysis techniques have already played an important role in this field, 
%the biological sequence data are quite far from being fully leveraged. 
%
On the other hand, the number of unlabeled protein sequences increases quite faster than that of the labeled ones, 
due to the large gap between low-cost, high-throughput sequencing methods and 
expensive yet time-consuming labeling approaches by time- and labor-intensive manual curation process~\citep{uniprot2019uniprot}. 
%Especially, this trend can be illustrated by the numbers presented 
%by UniProt~\citep{uniprot2019uniprot}, the biggest protein database in the world. %
%The Protein Data Bank (PDB) of experimentally determined structures is approaching $180,000$ entries\footnote{\url{https://www.rcsb.org/stats/growth/growth-released-structures}}, 
%while BFD\footnote{\url{https://bfd.mmseqs.com/}} aggregates billions of protein sequences.
%The Protein Data Bank (PDB) of experimentally determined structures is approaching $180,000$ entries, 
%while BFD aggregates billions of protein sequences.
%Nevertheless, the number of the sequences with known structures from PDB %, about 18,000, 
%is quite small compared to that of the sequences collected.
%%i.e., about 221 million sequences in UniRef and more than 4 billions sequences in BFD.
%In details, the Protein Data Bank (PDB) of experimentally determined structures is about 18,000, 
%while there are about 221 million sequences in UniRef and more than 4 billion sequences in BFD.
%
Understanding the protein sequences is vital to advance in biology, healthcare, and medicine.
The protein sequence representation has been exploited in various real applications, including
for dark proteome~\citep{perdigao2015unexpected}, where
the `dark' regions of proteins are never observed by experimental structure determination and inaccessible to homology modeling;
therapies, such as cancer diagnosis~\citep{vazquez2009modular} and antibody design~\citep{whitehead2012optimization};
function/property prediction for protein engineering, such as fitness prediction~\citep{hsu2021combining};
phylogenetic inference~\citep{hie2021evolutionary};
sequence mutations, such as learning mutational semantics~\citep{hie2020learning},
virus evolution and escape prediction~\citep{hie2021learning}, and
cancer diagnosis by mutation prediction~\citep{reva2011predicting,martelotto2014benchmarking},
to name a few.

%For the structure information, there are mainly two kinds of relations among the residues, i.e. local and global,
%while the global one is critical for the overall structure and function of the protein sequence.
%The global information of a sequence represents that whether the residues are spatially close to each other,
%the 3D structure sketch can be roughly determined when the global information is known.
%while the local information indicates how the protein folds in fine-grained details.
Co-evolutionary information represented by the inter-residue co-variation
is essential for protein sequences in terms of both structure and function
due to the evolution pressure from natural selection 
-- sequences with more stable structure and more adequate function are usually remained.
A representation that can capture this information from the sequences
is beneficial for understanding the molecular machinery of life.
This information is previously quantified by analyzing homologous sequences, 
\textit{a.k.a} multiple sequence alignment (MSA), for the target sequence,
where the homologs are retrieved from public protein sequence databases
by applying curated procedure and intensive computation with customized tools and hyper-parameters.
%
%\textbf{1)}
For protein function prediction, unsupervised models can be learned from these homologous sequences,
such as mutational effect prediction from sequence co-variation~\citep{hopf2017mutation} and
mutational landscape inference~\citep{figliuzzi2016coevolutionary}.
%and zero-shot function prediction~\citep{meier2021language}.
%
It is demonstrated that incorporating inter-residue dependencies using a pairwise model that
can power the predictions to agree more with the mutational 
experiment observations~\citep{mann2014fitness,boucher2016quantifying,riesselman2017deep}.
The reason is that, the function of the sequence is the combination effect of the residues 
other than the sum of the properties of each residue~\citep{hopf2017mutation}.
As the cost to quantify the co-effect of multiple residues is combinatorial
while modeling their interactions is critical, 
pairwise co-variation analysis becomes the best choice, meanwhile, 
\citet{figliuzzi2018pairwise} demonstrates that pairwise models are able to capture
collective residue variability.
%
%\textbf{2)}
For protein structure prediction, the key step is to predict
inter-residue contacts/distances, while the shared cornerstone of prediction is 
performing evolutionary coupling analysis, i.e. residue co-evolution analysis, 
on the constructed MSA for a target protein~\citep{ju2021copulanet}.
The underlying rational is that two residues which are spatially close in the three-dimensional structure
tend to co-evolve, which in turns can be exploited to estimate contacts/distances 
between residues~\citep{seemayer2014ccmpred,jones2018high}.
Although the contacts are usually rare in the residue pairs of the sequence, 
they are critical to rebuilding the 3D structure of the protein due to their roles of constraints on the sketch.

%Let the sequence be $X = (x_1, x_2, \ldots, x_N)$ and its MSA as $\mathcal{X} = \{ X_1, \ldots, X_K \}$.
%DCA can be formulated as follows:
%$$P(X \, | \, r, h) = \frac{1}{Z} \exp \left( \sum_{i=1}^{N-1} \sum_{j=i+1}^{N} r_{ij}(x_i, x_j) + \sum_{i=1}^{N} h_i(x_i)\right)$$
%For a protein family $\mathcal{X}$, 
%$$\Sigma_{ij}(a,b) = \frac{1}{K} \sum_{k=1}^{K} \left( x_{k, i}(a) - \bar{x}_{\cdot, i}(a) \right) \left( x_{k, j}(b) - \bar{x}_{\cdot, j}(b) \right)$$
%$$\Omega = \Sigma^{-1}$$
%%$$\max_{\Omega} \ \log P(\mathcal{X} \, | \, \Omega) - \lambda \, ||\Omega||_1$$
%%Assume $P$ to be Gaussian, the above optimization problem is equivalent to the following.
%%$$\max_{\Omega} \ \left( \log |\Omega| - tr(\Omega \, \hat{\Sigma}) \right) - \lambda \, ||\Omega||_1$$
%%Where $\hat{\Sigma}$ is the empirical covariance matrix calculated from $\mathcal{X}$.

%Thus, encoding co-evolutionary information into the embedding is essential for a useful protein sequence representation.
\textit{A natural question arises that, as both derived from unlabeled sequences,
can we directly capture this co-evolutionary information via sequence pre-training instead of extracting from MSA?}
To answer this question, we first discuss the key ingredient missed by conventional language models,
%for the co-evolutionary pre-training on sequences,
then we propose a pairwise masked language model for the protein sequence representation learning.
The key information to extract from the MSA (denoted as $\mathcal{X}$) is 
the statistics for $Q(x_i, x_j | \mathcal{X})$, %and $Q(x_i | \mathcal{M}), Q(x_j | \mathcal{M})$,
typically the co-variation for each pair, where $i, j$ are two indexes of the tokens/residues in the sequences. 
%The intuition is that, if the two residues are independent, these two probabilities will be equaled.
%Otherwise, their difference indicates the correlation between these two positions.
The intuition is that, if the two residues are independent, their co-variance will be close to zero,
otherwise, it indicates the co-evolutionary relation between these two positions.
\begin{table}[h]
	\centering
	\TableSize{}
	\caption{Example Residue Probabilities Conditioned on $X_{/\{A, B\}}$: Co-evolutionary \textit{vs.} Independent}
	\label{tab:example}
	\vspace{0.5em}
	\begin{tabular} {c | c c c | c}
		\hline
		 $A \not\perp B $ & $a_1$ & $a_2$ & $a_3$ & $P(B)$ \\
		 \hline
		 $b_1$ & $20\%$ & $20\%$ & - & $40\%$ \\
		 $b_2$ & $20\%$ & $10\%$ & - & $30\%$ \\
		 $b_3$ & - &  - & $30\%$ & $30\%$ \\
		 \hline
		 $P(A)$ & $40\%$ &  $30\%$ & $30\%$ & $100\%$ \\
		 \hline
	\end{tabular}
\qquad
	\begin{tabular} {c | c c c | c}
		\hline
		$A \perp B $ & $a_1$ & $a_2$ & $a_3$ & $P(B)$ \\
		\hline
		$b_1$ & $16\%$ & $12\%$ & $12\%$ & $40\%$ \\
		$b_2$ & $12\%$ & $9\%$ &  $9\%$ & $30\%$ \\
		$b_3$ & $12\%$ & $9\%$ & $$9\%$$ & $30\%$ \\
		\hline
		$P(A)$ & $40\%$ &  $30\%$ & $30\%$ & $100\%$ \\
		\hline
	\end{tabular}
\end{table}
While for the conventional masked language model,
although they scan through all the same sequences as MSA-based methods 
and the losses for all the residues in the masked tokens will be back-propagated, 
they are processed in an accumulative way within the batches, i.e. independently updating the weights. 
This means that $P(x_i | X_{/M})$ is modeled by the LMs.
Table~\ref{tab:example} presents an example for this kind of method, originating from NLP but
sub-optimal for protein sequences
due to the critical and much frequent co-evolution relationship among the residues.
Here, we argue that for the protein sequences $P(x_i, x_j | X_{/M}) \neq P(x_i | X_{/M}) \cdot P(x_j | X_{/M})$.
Auto-regressive methods, i.e. generative pre-training (GPT), 
where the latter tokens are predicted by conditioning on the previous ones from the same direction, 
still having the similar issue.
Inspired by this,
we design a pairwise masked language model and pre-train the model with a pairwise loss calculated from the protein sequences directly.
Following the masked language model where the tokens of each sequence are picked at a probability and 
then be masked, replaced, or kept, our proposed model takes the masked sequences as input and predicts these original tokens,
especially, the label for each masked token pair.

To examine the capability of the model to capture the co-evolutionary information, 
we select protein contact prediction as our main downstream task
due to its well-established relationship to the co-evolutionary information and 
the plentiful data that are publicly available and well measured by empirical experiment.
Here, protein contact prediction is a binary classification task for amino acid residue pairs,
a residue pair is called a contact if their distance is less than or equal to a distance threshold, 
%typically $8 \ \text{\AA{}}$\footnote{$1 \ \text{\AA{}} = 10^{-10} \ \text{m}$}.
typically $8 \ \text{\AA{}}$ (i.e., $8 \times 10^{-10} \ \text{m}$). % ($8 \times 10^{-10}$ $\mathbf{m}$)
The result shows that this method can significantly improve the learned representation on the contact prediction task,
it even surpasses the MSA baseline when pre-trained on another sequence subset UR50
with the sequence-only input on the TAPE benchmark.
Meanwhile, an additional experimental evaluation for secondary structure prediction illustrates 
that the proposed model does not hurt performance, 
while significant improvements are observed in some other settings for remote homology prediction.

\section{Related Work}
\label{related}

\paragraph*{Evolutionary Coupling Analysis}

Co-evolution information is closely correlated to the contacts,
due to the rational that two residues are likely co-evolving when they are spatially close.
To this end, techniques such as mutual information (MI)~\citep{gloor2005mutual} are exploited to quantify this feature. 
Later than MI, direct coupling analysis (DCA) proves to be more accurate and be widely adopted,
for example, EVfold~\citep{sheridan2015evfold}, PSICOV~\citep{jones2012psicov}, 
GREMLIN~\citep{kamisetty2013assessing}, plmDCA~\citep{ekeberg2013improved}, CCMpred~\citep{seemayer2014ccmpred} are all built on DCA.
%
%DCA is an umbrella term comprising several methods for analyzing sequence data in computational biology. 
For protein sequences, the common idea of these methods is 
to use statistical modeling to quantify the strength of the direct relationship 
between two positions of a protein sequence, excluding effects from other positions.
A variety of DCA methods are further proposed to generate the direct couplings of residues 
%by applying the inverse of covariance matrix or Potts model,
by fitting Potts models~\citep{ekeberg2013improved} or precision matrix~\citep{jones2012psicov} to MSAs, 
e.g. mean-field DCA~\citep{morcos2011direct}, sparse inverse covariance~\citep{jones2012psicov} and pseudo-likelihood 
maximization~\citep{ekeberg2013improved, balakrishnan2011learning,  seemayer2014ccmpred}.
%
%Many statistical modeling methods based on direct coupling analysis~(DCA) have been proposed
%to quantify the strength of the direct relation between the residue pairs of a protein sequence
%while excluding effects from other residues
%by fitting Potts models~\citep{ekeberg2013improved} or precision matrix~\citep{jones2012psicov} to MSAs, 
%such as mean-field DCA~\citep{morcos2011direct}, 
%sparse inverse covariance~\citep{jones2012psicov} and pseudo-likelihood 
%maximization~\citep{ekeberg2013improved, balakrishnan2011learning,  seemayer2014ccmpred}.
%These methods further exploit dedicated scores based on DCA for contact prediction.

By taking DCA-derived scores as features, 
deep neural networks based supervised methods significantly outperform 
the unsupervised methods~\citep{senior2020improved, jones2018high, wang2017accurate, xu2019distance, yang2020improved}. 
Most neural network models, including AlphaFold~\citep{alquraishi2019alphafold} 
and RaptorX~\citep{xu2019distance}, rely on this feature.
However, due to the considerable information loss 
after transforming MSAs into hand-crafted features,
supervised models, such as CopulaNet~\citep{ju2021copulanet} and AlphaFold2~\citep{jumper2021highly}, 
are proposed to directly build on the raw MSA.
The superior performance over the baselines demonstrates that residue co-evolution information can be
mined from the raw sequences by the model.
A noticeable drawback of the MSA based methods is that, when the MSA quality (i.e. the number of the homologous sequences) is low
for a target sequence, the model performance drops a lot.

%By taking DCA-derived scores as features, 
%deep neural networks based supervised methods significantly outperform 
%the unsupervised methods~\cite{senior2020improved, jones2018high, wang2017accurate, xu2019distance, yang2020improved}. 
%However, the information lost by the DCA-based features from the sequences is still not recoverable. 
%% site-independent & pairwise-interactions
%%
%To mitigate this issue, several models are proposed to learn residue co-evolution information 
%directly from the sequences in the MSAs~\cite{ju2021copulanet, kandathil2020deep, mirabello2019rawmsa, jumper2020high}. 
%Among them, CopulaNet~\cite{ju2021copulanet}, the SOTA of the CASP13 benchmark, 
%derives coevolutionary features differentially by aggregating the learned residue representations from the sequences.
%However, although they are capable of modeling the high-order interactions among the multiple residues within single sequence, 
%the global information carried by the MSAs is ignored because they still model the protein sequences independently.
%AlphaFold2~\cite{jumper2020high} is claimed to be modeling the full MSAs, 
%achieving an amazing performance on the CASP14 structure prediction task. 
%Although its performance on the structure prediction task is remarkable, 
%they did not participate in the CASP14 PCP task meanwhile no further details of their methods are publicly available.

\paragraph*{Pre-Training Methods} 
%Following the \textit{pre-train and fine-tune} paradigm,
%pre-trained language models are adapted to representation learning for protein 
%sequences from the unlabeled data~\cite{elnaggar2020prottrans, rao2019evaluating, rao2020transformer, rao2021msa, rives2021biological, sturmfels2020profile}. Many of them take contact prediction as an important downstream task to validate their performance.
%While these methods show another solution to this task, they are still at an early stage
%thus cannot achieve comparable performance to the SOTA approaches currently. 
%%
%To further improve the performance, a pre-trained language model named {MSA Transformer} 
%is proposed to learn a better MSA representation directly~\cite{rao2021msa}.
%This model shows the possibility of modeling the co-evolution information from MSAs,
%but it still cannot well extract the co-evolution information
%due to the non-homologous subsequences inevitably introduced by the generated MSAs.

Following the \textit{pre-train} and \textit{fine-tune} paradigm~\citep{peters2018deep,kenton2019bert,liu2019roberta,radford2018improving}, 
various pre-training methods are proposed recently to learn better representations for protein sequences.
TAPE~\citep{rao2019evaluating} is built as a benchmark to evaluate the protein sequence pre-training method.
It demonstrates the performance can be improved by pre-training compared to one-hot representation, however, also
indicates that the performance of the pre-training based models on pure sequences still lags behind the alignment-based method in the downstream tasks.

Among pre-training methods, RNNs are exploited as the pre-training model.
\cite{bepler2019learning} use two layers of Bi-LSTM as an extraction part of the original protein sequences 
by applying next token prediction for pre-training language modeling. 
UniRep~\citep{alley2019unified} uses a similar training scheme as \cite{bepler2019learning}, then use evo-turning (evolutionary fine-tuning) 
to address the importance of evolutionary information for the protein embedding.
UDSMProt~\citep{strodthoff2020udsmprot} relies on an AWD-LSTM language model, 
which is a three-layer LSTM regularized by different kinds of dropouts.

A variety of techniques based on Transformer~\citep{vaswani2017attention} are applied to build dedicated models for proteins.
ESM~\citep{rives2021biological} demonstrates that Transformer models can outperform RNN based models,
it also illustrates that the protein sequence diversity and model size 
have significant impacts on the pre-trained encoder performance.
PRoBERTa~\citep{nambiar2020transforming} follows RoBERTa to pre-train the model with Byte Pair Encoding
and other optimization techniques.
\cite{lu2020self} applies contrastive learning by mutual information maximization to the pre-training procedure,
MSA Transformer~\citep{sturmfels2020profile} learns the representation on the MSAs for a protein sequence,
however, it relies on expensive database searches to generate the required MSAs for each sequence.

Large-scale models are explored due to the fact that protein sequence datasets are large in size and complex in interaction.
ProtTrans~\citep{elnaggar2021prottrans} trains the protein LMs (pLMs) on the Summit supercomputer using
more than $4$ thousand GPUs and TPU Pod up-to $1024$ cores,
the most informative embeddings even outperform 
state-of-the-art models without multiple sequence alignments (MSAs) for the secondary structure prediction.
%dimensionality reduction revealed that the raw pLM-embeddings from unlabeled data
%captured some biophysical features of protein sequences;
%The models are evaluated on the 3-state protein secondary structure (per-residue task) and protein sub-cellular location prediction (per-protein task).
%For secondary structure, the most informative embeddings for the first time outperformed the
%state-of-the-art without multiple sequence alignments (MSAs) or evolutionary information thereby bypassing expensive database searches. 
%Taken together, the results implied that pLMs learned some of the grammar of the language of life.
\cite{xiao2021modeling} also demonstrate that sequence evolution information can be accurately captured 
by a large-scale model from pre-training, up to $3$ billion parameters pre-trained on a $480$ GPUs cluster.

Additional supervised labels are exploited for protein sequence pre-training.
PLUS~\citep{min2019pre} tries to model the protein sequence with masked language model together with an auxiliary task, 
i.e. same family prediction, in their work. 
\cite{sturmfels2020profile} add a pre-training task named profile prediction for pre-training.

Generative pre-training is also exploited for protein engineering, for example,
ProGen~\citep{madani2020progen} is a generative model conditioned on taxonomic information
as an unsupervised sequence
generation problem in order to leverage the exponentially growing set of proteins that lack costly, structural annotations.

The outputs of the pre-trained language models can be used in different ways.
For example, \cite{vig2020bertology}, \cite{rao2020transformer}, and \cite{bhattacharya2020single} demonstrate that 
the attention weights from the pre-trained models have a strong correlation with the residue contacts.
%The language models can also be used for the protein sequence mutations. For example, 
\cite{hie2021evolutionary} analyze the correlation between 
the pseudo likelihood of the mutated sequences predicted by the language model and the evolution of the mutations,
the results suggests that pre-trained language models on protein sequences are able to predict evolutionary order at different timescales.
\cite{hie2021learning,hie2020learning} apply pre-trained protein language models to predict mutational effects and virus escape.

\section{Method}

Given the sequence data $\mathcal{D} = \{ X \}$, where $X = (x_1, x_2, \ldots, x_N) $, 
language models can be built on the sequences for pre-training.

\subsection{Pairwise Masked Language Model}

The loss function of vanilla Masked Language Model (MLM) can be formulated as follows:

$$\mathcal{L}_{mlm} = \mathbb{E}_{X \sim \mathcal{D}} \, \left( \mathbb{E}_{M} \sum_{i \, \in \, M} \left( - \log P_\theta(x_i \, | \, X_{/M}) \right) \right)$$

where $\mathcal{D}$ is the sequence set, $X$ is a sequence in $\mathcal{D}$,
$X_{/M}$ represents the masked sequence of $X$ where the masked token indices are in $M$, 
$x_i$ stands for the $i$-th token in the sequence $X$,
and $\theta$ denotes the encoder parameters.

In this paper, we propose a novel Pairwise Masked Language Model (PMLM) whose loss function can be written as:

$$\mathcal{L}_{pmlm} = \mathbb{E}_{X \sim \mathcal{D}} \, \left( \mathbb{E}_{M} \sum_{i, j \, \in \, M, \, i \neq j} \left( - \log P_\theta(x_i, x_j \, | \, X_{/M}) \right) \right)$$

The combined loss for both MLM and PMLM can be used for pre-training as:

$$\mathcal{L} = \mathcal{L}_{mlm}  + \lambda \cdot \mathcal{L}_{pmlm} $$ 

% Here, we argue that 
% $$ P_\theta(x_i, x_j \, | \, X_{/M}) \neq P_\theta(x_i\, | \, X_{/M}) \cdot P_\theta(x_j \, | \, X_{/M}) \quad \textit{s.t.} \quad i, j \, \in \, M, \, i \neq j $$

where $\lambda$ is a weight coefficient to balance two losses.
In this paper, $\lambda$ is set to $1$ during pre-training the PMLM models.
As we can see, all the MLM and PMLM labels for pre-training are from the sequences themselves, thus still self-supervised.
For multiple rounds of updates, if $i$-th and $j$-th positions are independent, 
$P_\theta(x_i, x_j | X_{/M})$ degenerates to $P_\theta(x_i | X_{/M}) \cdot P_\theta(x_j | X_{/M})$.
When they are co-evolutionary, the model learns a different distribution from the independent case.

\subsection{Model Architecture}

\begin{figure}[h]
	\centering
	\includegraphics[width=0.85\textwidth]{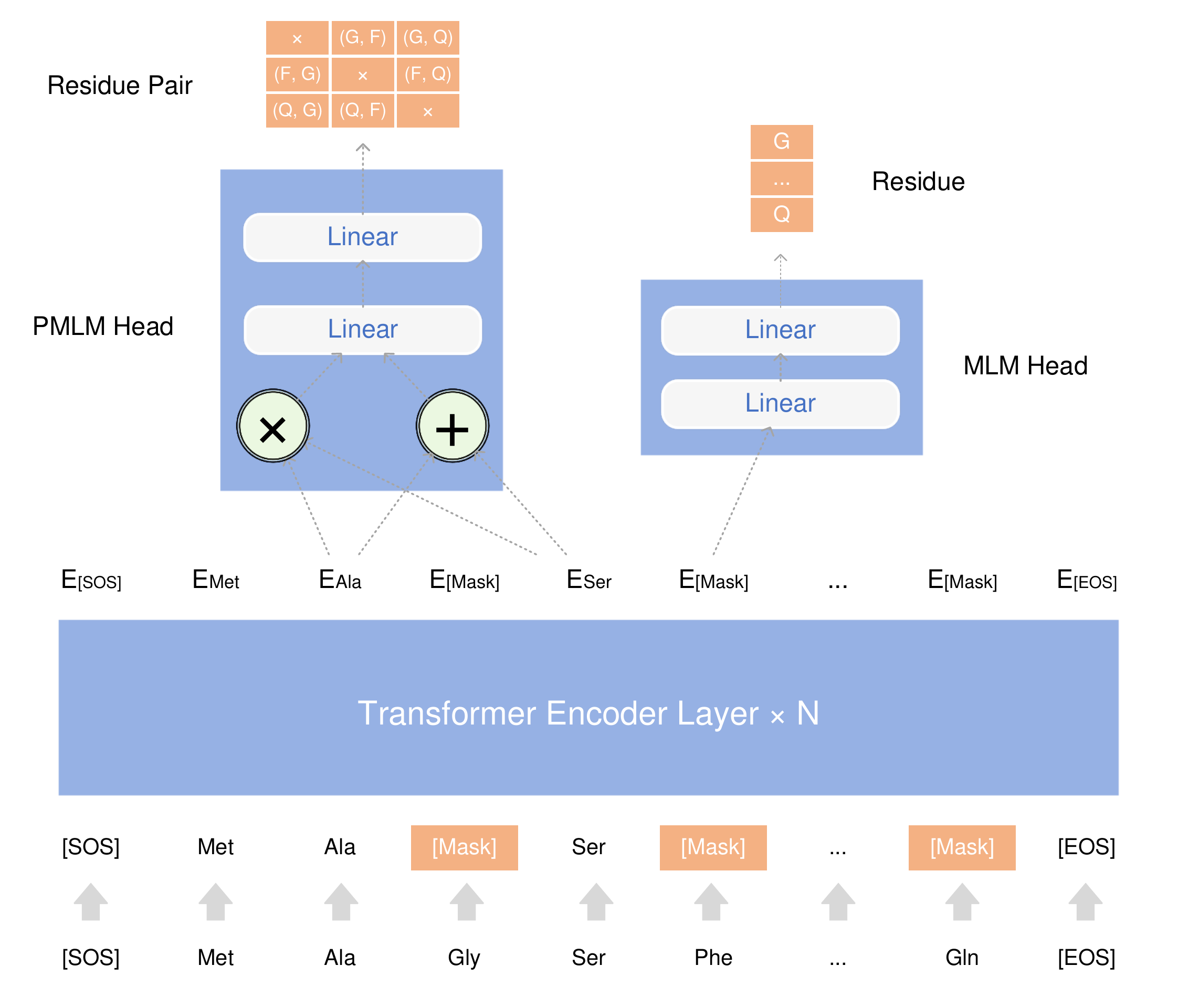}
	\caption{Overview of the Pairwise Masked Language Model}
	\label{fig:pmlm}
\end{figure}

The goal of pre-training is to build a good protein sequence encoder that generates better representation.
The pre-training model mainly consists of a protein sequence encoder and two prediction heads,
i.e., a \textit{token prediction head} and a \textit{pair prediction head}. %on top of the sequence encoder.
The sequence encoder is built on stacked Transformer encoder layers.
The overview of our model is illustrated in Figure~\ref{fig:pmlm}.
Transformer is believed to be a powerful tool for modeling sequence data and has been applied to various tasks, 
including natural language understanding, question answering, computer vision, and so on.
Thus we exploit a Transformer encoder as the sequence encoder of our model.
The sequence encoder takes raw protein sequences as input and converts them into their vector representations. 
The model was trained on protein sequences using both 
masked token prediction (MLM) and masked pair prediction (PMLM) for protein language modeling.
Each prediction head is a MLP, i.e. two-layer fully-connected (FC) neural network, 
where the output is mapped into the vocabulary via softmax,
i.e. $20$ amino acid residues (denoted as $V_{res}$) for the token prediction head 
and $400$ amino acid pairs (denoted as $V_{pair}$) for the pair prediction head.

\subsection{Masked Pair Prediction}

The masked language model lets the model to reconstruct the masked tokens conditional on the other tokens in the sequences.
Predicting masked pairs follows the same idea of the masked token prediction, however, with losses for the pairs.
A two-layer FC neural network is exploited for the prediction of masked pairs,
whereas another one for that of masked tokens.
During pre-training, the dot and difference of the vectors for the residue pair are concatenated
before feeding into the pair prediction head.
Then, the output is mapped into a vector of dimension $|V_{pair}|$ by softmax.
This prediction is finally compared with the pair label from the sequence itself.

\begin{figure}[h]
	\centering
	\includegraphics[width=0.7\textwidth]{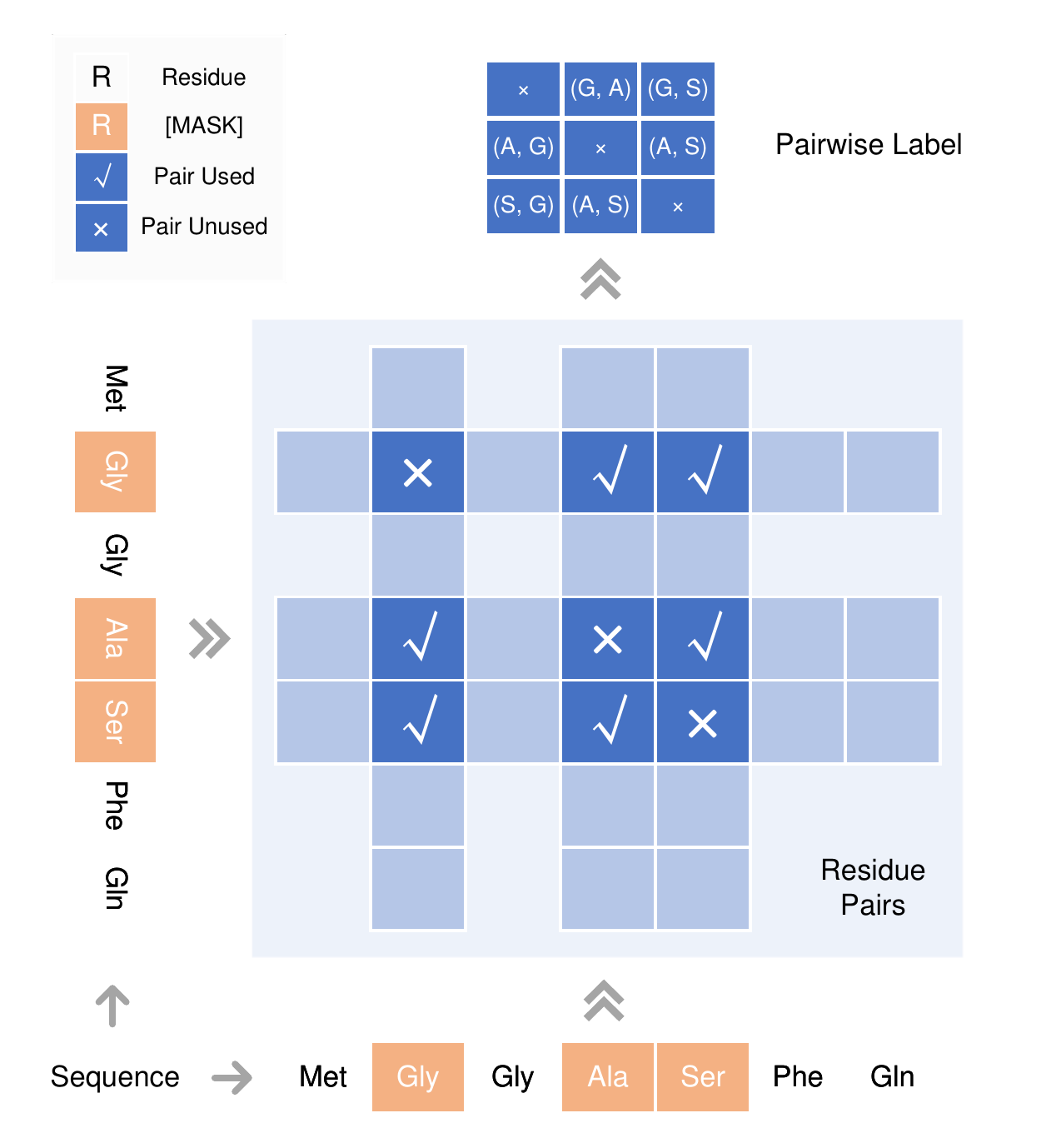}
	\caption{Pairwise Label Construction from the Masked Sequence}
	\label{fig:pairwise-label}
\end{figure}

It is not trivial to generate pair labels from multiple masked tokens.
The pairwise label construction process is demonstrated in Figure~\ref{fig:pairwise-label}.
For each pair $x_i, x_j$ for the masked token where $i \neq j$, 
a pairwise label is generated as $w_{ij} = (x_i, x_j)$,
where $x_i, x_j \in V_{res}$ and $w_{ij} \in V_{pair}$.
Obviously, we have $|V_{pair}| = |V_{res}|^2$ for $V_{res}$ and $V_{pair}$.
The pairs for $x_i, x_j$ whose $i=j$ are ignored since they are already a part of the MLM loss.
Notably, $i \neq j$ does not necessarily mean $x_i \neq x_j$ since two different positions may be the same residue.
For each valid pair, the pair prediction head generates the probability over $V_{pair}$. 
For a masked sequence $X_{/M}$, the size of the pairwise labels is $|M|^2 - |M|$
while each label is a one-hot encoding for $V_{pair}$.

Intuitively, the masked pair loss is consistent with the masked token loss. For example, 
the loss can be treated as a combined loss for the two masked tokens if the two positions within the pair are independent, 
otherwise this masked pair loss provides extra information of the sequence besides each position.

\subsection{Fine-tuning for Contact Prediction}

The generated encoder is further fine-tuned on a given supervised dataset for contact prediction.
As the prevalent approach does, different neural networks can be built on top of the pre-trained model for further fine-tuning,
whereas, a simple FC layer followed by a softmax operator can be applied to evaluate the representation.
For the residue pairs task,  e.g. contact prediction, 
the fine-tuning network is built on top of the first FC layer of the pair prediction head,
other tasks are fine-tuned on the encoder outputs.

%[TODO]

\section{Experiment Evaluation}
\label{sec:experiment}

\subsection{Experiment Setup}

To evaluate the performance of the pairwise masked language model, 
several models with different settings are pre-trained on two prevalent datasets, i.e. Pfam~\citep{el2019pfam} and UR50 (release 2018\_03).
Following the RoBERTa-base setting, 
the hidden size, feed forward dimension, number of encoder layers, and attention heads of the base models are set
as $768$, $3072$, $12$, $12$ respectively (denoted as MLM-base for MLM and PMLM-base for PMLM).
A larger model named PMLM-large is pre-trained 
with the same setting except using $34$ as the number of encoder layers.
Moreover, the largest model with a hidden size of $1280$ 
and a number of encoder layers of $36$ is also pre-trained on UR50,
denoted as PMLM-xl. The implementation is optimized to speed up the training procedure, e.g.,
the pairwise loss is applied alone by enabling the diagonal elements and disabling the token prediction head.
MLM-base and PMLM-base are pre-trained with maximum length $512$,
while PMLM-large is pre-trained with a maximum length of $1024$.
The positional encoding of both models are non-learnable.
MLM-base and PMLM-base are pre-trained using the Adam optimizer $(0.9,0.98)$ 
with peak learning rate $0.0003$ and clip norm $1.0$,
the learning rate scheduler is a polynomial decay scheduler where the learning rate is decreased linearly
after warming up by $20,000$ steps to the peak.
The hyper-parameters are almost the same for pre-training PMLM-large
except that the peak learning rate is set to $0.0001$.
MLM-base, PMLM-base, and PMLM-large are pre-trained on $24$ Tesla V100 GPU cards,
about three weeks for MLM-base/PMLM-base and about seven weeks for PMLM-large.
PMLM-xl is pre-trained on $16$ Tesla V100 GPU cards for more than two weeks.

%
%The model settings are listed in Table~\ref{tab:hyperparameters}.

%\begin{table}[h!]
%	\centering
%	\caption{Model Settings}
%	\label{tab:hyperparameters}
%	\vspace{0.5em}
%	\begin{tabular}{c|c|c|c|c}
%		\hline
%		\textsc{Model} & \textsc{Hidden Size} & \textsc{FFN Dim} & \textsc{Encoder Layers} & \textsc{Attention Heads} \\ 
%		\hline\hline
%		MLM-base &  768  &  3,072 &  12 & 12 \\ 
%		PMLM-base &  768  &  3,072 &  12 & 12 \\ 
%		PMLM-large &  768  &  3,072 &  34 & 12 \\ 
%		\hline
%	\end{tabular}
%\end{table}

Two groups of experiments are conducted to examine the generated model on 
the TAPE contact prediction test set (denoted as TAPE-Contact) and 
the RaptorX contact prediction test set (denoted as RaptorX-Contact).
The models are all evaluated on or fine-tuned from the checkpoint of the sixtieth epoch,
except PMLM-base on UR50, which is fine-tuned from that of the ninetieth epoch.
For TAPE-Contact, we use a shallow decoder as TAPE does, i.e.
simply use an outer-dot and outer-difference of each pair followed by a linear projection with softmax, 
for the sequence representation to evaluate the representation.
To see the potential of the representations, RaptorX-Contact is evaluated on a more expressive decoder, 
i.e., a ResNet with stacked convolution blocks, on top of the pre-trained encoder as ESM does,
however, the convolution blocks are not dilated.
The fine-tuning process for each model on each dataset is finished on one Tesla V100 GPU card.

\subsection{Pre-training Validation}

\begin{figure}[h]
	\centering
	\begin{subfigure}[b]{0.48\textwidth}
		\centering
		\includegraphics[width=\textwidth]{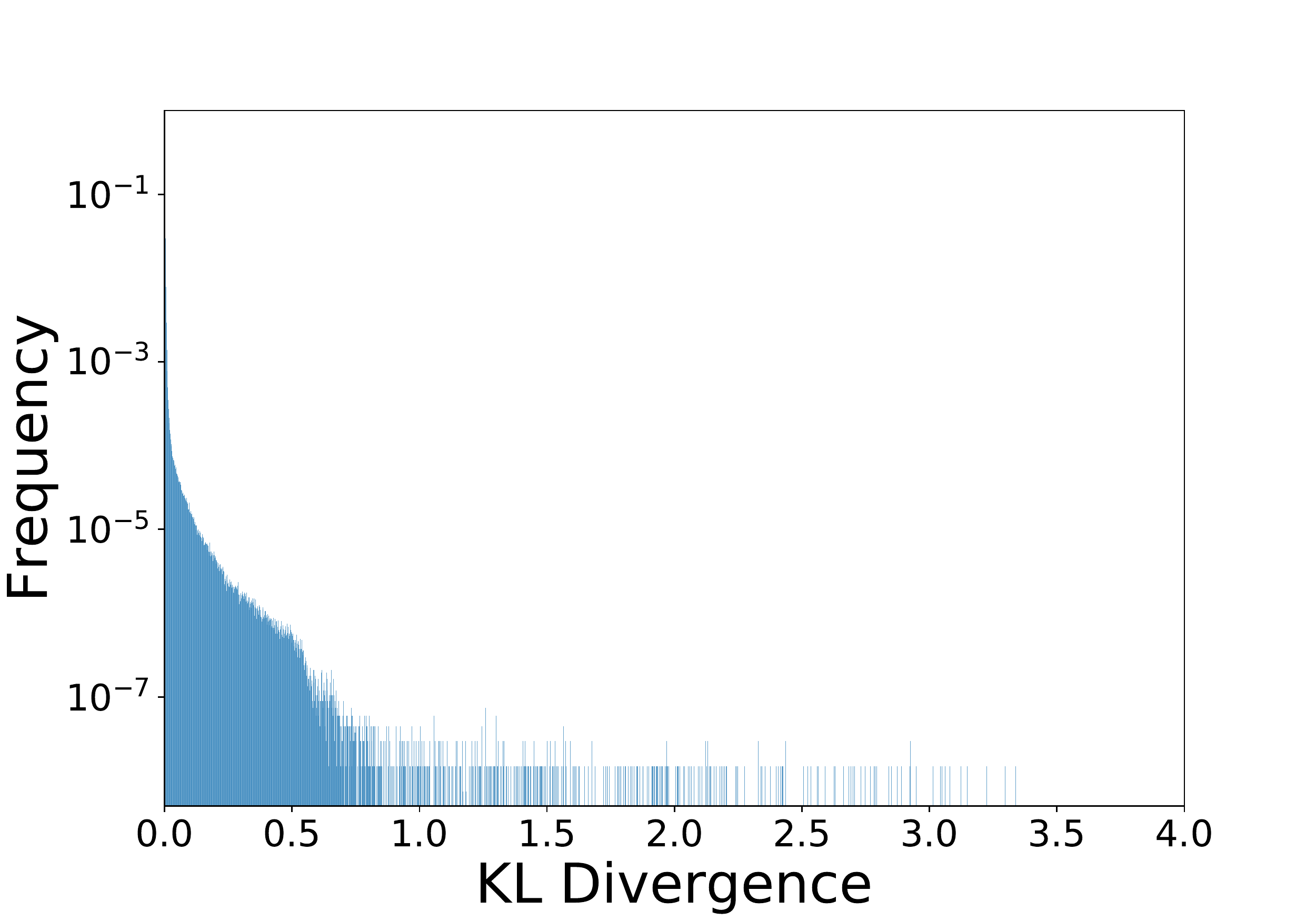}
%		\caption{$y=x$}
%		\label{fig:y equals x}
	\end{subfigure}
	\hfill
	\begin{subfigure}[b]{0.48\textwidth}
		\centering
		\includegraphics[width=\textwidth]{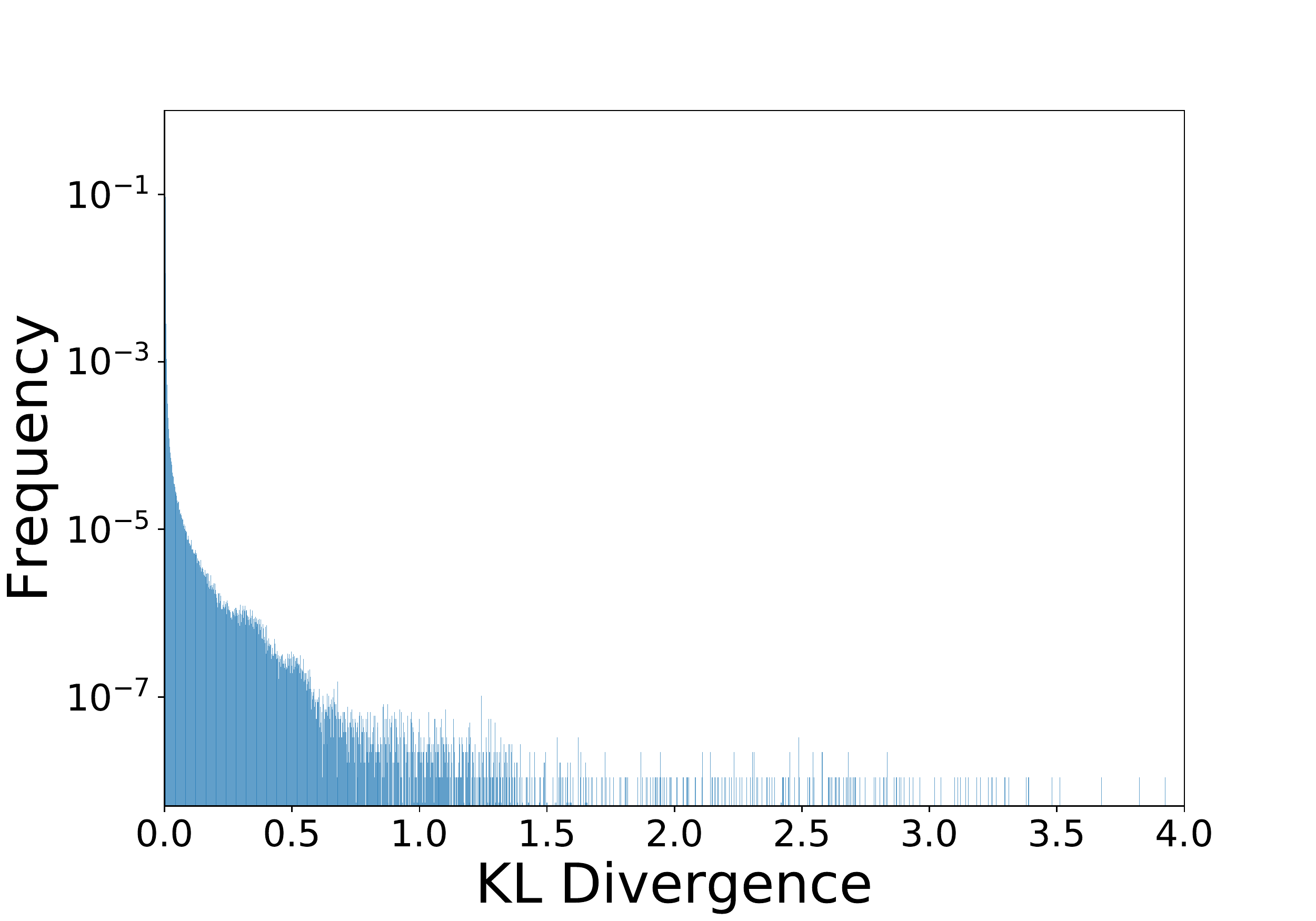}
%		\caption{$y=3sinx$}
%		\label{fig:three sin x}
	\end{subfigure}
	\caption{KL Divergence of PMLM-large on Pfam Valid (Left) and UR50 Valid (Right)}
	\label{fig:kldiv}
\end{figure}

The two heads of the PMLM model output $P(x_i, x_j)$ and $P(x_i), P(x_j)$ separately, this raises another question
that if they have already learned the difference between the independent and dependent positions of the sequence.
To answer this question, we aggregate the validation losses and accuracy scores of pre-training
on the Pfam and UR50 sequence datasets, as shown in Table~\ref{tab:pretraining-scores}.
As we can see, even the fine-tuned performance of PMLM-base is better than that of MLM-base,
the loss $\mathcal{L}_{mlm}$ of PMLM-base is slightly higher than that of MLM-base.
We further use $\Delta\textsc{Acc}=\textsc{Acc}_{pmlm} - \textsc{Acc}_{mlm}^2$ to quantify the difference.
For PMLM-base and PMLM-large on Pfam and UR50, 
$\Delta\textsc{Acc}$ is not negligible as an expectation,
which indicates the difference of the correct probability for the most likely residue on each position.
The joint probability is quite different to the multiplication of the individual probability,
illustrating that PMLM is able to capture the co-evolutionary information via pre-training on pure sequences.
This observation can be further validated by the histograms of 
$KL \left( P(x_i) \cdot P(x_j) \, || \,  P(x_i, x_j) \right)$
of the PMLM-large model on the Pfam and UR50 validation sequences, which is shown in Figure~\ref{fig:kldiv},
each $(x_i, x_j)$ pair of the sequence is masked when predicting the probabilities.

\begin{table}[h!]
	\centering
	\TableSize{}
	\caption{Validation Losses and Accuracy Scores of Pre-training} % (@Epoch 60)
	\label{tab:pretraining-scores}
	\vspace{0.5em}
	\begin{tabular}{c|c|c|c|c|c|c|c}
		\hline
		\textsc{Model} & \textsc{Data} & \textsc{$\mathcal{L}_{mlm}$} & \textsc{$\mathcal{L}_{pmlm}$} & \textsc{Acc$_{mlm}$} & \textsc{Acc$_{pmlm}$} & \textsc{$\Delta$Acc} & \textsc{$\Delta$Acc / Acc$_{pmlm}$ }\\ 
		\hline\hline
		%%% loss / ln2 =
		%		MLM-base  & Pfam  &  2.37688 &  - & 48.0\% & - \\ 
		%		PMLM-base  & Pfam &  2.41342 &  4.82526 & 47.1\% & 22.4\% \\ 
		%		\hline
		%		PMLM-base & UR50  & 3.28699 &  6.57253 & 31.8\% & 10.9\%  \\ 
		%		PMLM-large & UR50  & 3.13205 &  6.26404 & 35.1\% & 13.2\%  \\ 
		MLM-base  & Pfam  &  1.6475 &  - & 48.0\% & - & - & - \\ 
		PMLM-base  & Pfam &  1.6728 &  3.3446 & 47.1\% & 22.4\% & 0.22\% & 1.0\%\\ 
		\hline
		MLM-base & UR50 & 2.2725 & - & 32.0\% & - & - & - \\ 
		PMLM-base & UR50  & 2.2784 &  4.5557 & 31.8\% & 10.9\% &  0.79\% & 7.2\% \\ 
		\hline
		PMLM-large & UR50  & 2.1710 &  4.3419 & 35.1\% & 13.2\% &  0.88\% & 6.7\% \\ 
		PMLM-xl & UR50  &  - &  3.9746  &  37.3\%  & 16.9\%  &  2.95\%  & 17.5\% \,  \\ 
		\hline
	\end{tabular}
\end{table}

%\begin{figure}[tbp]
%	\center
%	\includegraphics[width=\textwidth]{figures/pfam-kldiv}
%	\caption{KL Divergence Scores of PMLM-large on Pfam (Left) and UR50 (Right)}
%	\label{fig:kldiv}
%\end{figure}

\subsection{Comparison on TAPE-Contact}

For a fair comparison, the models are evaluated on the TAPE-Contact with the same contact prediction head,
which is an FC layer with a $0.5$ dropout followed by a softmax operator.
Medium and long range precision at $L/5$ is used for evaluation,
where medium and long range means the there are at least $12$ residues between the pair to test.
With precision at $L/5$, the top $L/5$ contact predictions are compared to the ground-truth,
where $L$ is the length of the input sequence.

To better control the setting, a Transformer encoder for MLM (denoted as MLM-base) as well as 
a Transformer encoder (denoted as PMLM-base)
are pre-trained on the Pfam and UR50 datasets.
The hyper-parameters for these models resemble that of the RoBERTa-base for NLP,
i.e., hidden size as $768$, feed forward dimension as $3072$, and number of layers as $12$.
To further evaluate the effect of PMLM, we also pre-train a deeper Transformer encoder 
with the number of layers increased to $34$, denoted as PMLM-large.

The fine-tuned performance of the models are compared in Table~\ref{tab:tape-mlrange}.
The performance of TAPE transformer and MSA baseline are reported in the paper.
As we can see, the performance is increased from $36.0$ for TAPE to $37.3$ for MLM-base.
%due to the larger model size.
Comparing the two models which is close in model size pre-trained on Pfam, i.e. MLM-base and PMLM-base, 
a huge performance gain over $9\%$ is witnessed, demonstrating the positive effect of the PMLM for pre-training.

Notably, PMLM-xl pre-trained on UR50 can even outperform the MSA baseline (about $8\%$),
%when fine-tuned with an additional 32-layer ResNet atop
%(each skipped layer consists of two sequential blocks of batch norm, ReLU activation, and 1-D convolution with kernel size 3),
although UR50 is a small subset of UniParc + Metagenome,
which shields the light for the single sequence pre-training methods to exceed the MSA based methods.
%To our knowledge, this is the first result reported to surpass the MSA baseline on this benchmark.

\begin{table}[h!]
	\centering
	\TableSize{}
	\caption{Medium-/Long- Range Precision@$L/5$ on TAPE Contact Prediction Test Set. 
		The predictions are generated from a linear projection of the encoder outputs during fine-tuning.
%		except that of PMLM-xl + ResNet, which leverages a ResNet with stacked convolution blocks.
		UR50 is a small subset of UniParc + Metagenome.
%		$^*$ The maximum sequence length is set to 768 due to the GPU memory usage.
	}
	\label{tab:tape-mlrange}
	\vspace{0.5em}
	\begin{tabular}{c|c|c|c|c}
		\hline
		\textsc{Model} & \textsc{\#Params} & \textsc{Data} & \textsc{Pre-train Task} & \textsc{Result}\\ 
		\hline\hline
		TAPE Transformer & 38M & Pfam & MLM & 36.0\\ 
		MLM-base & 85M & Pfam & MLM & 37.3\\ 
		% PMLM-base & 85M & Pfam & MLM + PMLM & 46.6\\ 
		% PMLM-base$^+$ & 87M & Pfam & MLM + PMLM & 47.2 \\  
		PMLM-base (ours) & 87M & Pfam & MLM + PMLM & 47.2 \\ 
		\hline
%		MLM-base & 85M & UR50 & MLM & 56.5 \\ 
%		PMLM-base (60)  & 85M & UR50 & MLM + PMLM & ?\\ 
%%		PMLM-base$^+$ (60) & 87M & UR50 & MLM + PMLM & ? (maxlen768) \\ 
%		PMLM-base (90)  & 85M & UR50 & MLM + PMLM & 57.7\\ 
%%		PMLM-base$^+$ (90) & 87M & UR50 & MLM + PMLM & 55.7 (maxlen768) \\ 
		PMLM-base (ours) & 87M & UR50 & MLM + PMLM & 57.7\\ 
		% PMLM-base$^+$ & 87M & UR50 & MLM + PMLM & 55.7 (maxlen768) \\ 
		% PMLM-large & 248M & UR50  & MLM + PMLM & 64.3\\ 
		% PMLM-large$^+$ & 250M & UR50  & MLM + PMLM & \textbf{\ \ 66.7}$^*$ \\ 
%		PMLM-large & 250M & UR50  & MLM + PMLM & \textbf{\ \ 66.7}$^*$ \\ 
		PMLM-large (ours) & 250M & UR50  & MLM + PMLM & 66.7 \\ 
%		PMLM-large + ResNet (ours) & 253M & UR50  & MLM + PMLM & 68.0 \\ 
		PMLM-xl (ours) & 713M & UR50  & MLM + PMLM & \textbf{71.7} \\ 
%		PMLM-xl + ResNet (ours) & 715M & UR50  & MLM + PMLM & \textbf{73.2} \\ 
		\hline
		MSA baseline & - & UniParc + Metagenome & - & 64.0 \\ 
		\hline
	\end{tabular}
\end{table}

\subsection{Comparison on RaptorX-Contact}

%MLM-base and PMLM-base are pre-trained with maximum length $512$,
%the positional encoding of both models are all non-learnable.

The models are evaluated on the RaptorX-Contact with the same contact prediction module,
however, with different contact prediction modules, 
customized networks may be optimal for different pre-trained encoder.

%As shown in Table~\ref{tab:esm-lrange},
%PMLM-base (+ Linear) outperforms MLM-base (+ Linear) by more than $5\%$ given that
%they are fine-tuned with all linear prediction head for the task,
%demonstrating the effect of PMLM for pre-training again.
%When comparing the models pre-trained on UR50,
%PMLM-base  (+ Linear) can even outperform ESM Transformer-12 (+ ResNet-32), 
%even that a more capable prediction head is exploited,
%showing the superior performance of the PMLM encoder. 

As shown in Table~\ref{tab:esm-lrange},
PMLM-base outperforms ESM Transformer-12 by about $4\%$ given that
their model sizes are close,
demonstrating the effect of PMLM for pre-training again.
When comparing the larger models of a close model size pre-trained,
PMLM-xl significantly outperforms ESM models (i.e., more than $9\%$ for ESM Transformer-34 
and $3\%$ for ESM-1b)
even that a systematic hyper-parameter searching on a 100M model 
is conducted to generate the hyper-parameters for ESM-1b,
which is not performed for PMLM-xl,
showing the superior performance of the PMLM encoder. 

\begin{table}[h!]
	\centering
	\TableSize{}
	\caption{Long-Range Precision@$L$ on RaptorX Contact Prediction Test Set.
		A ResNet with stacked convolution blocks is built on top of each pre-trained encoder during fine-tuning.
%		The hyper-parameter of ESM-1b is generated from a systematic optimization of model hyper-parameters,
%		while those of PMLM model is not.
		The performance numbers of the models except PMLM are reported by \citet{rives2021biological}.
%		$^*$The performance result of PMLM-xl is evaluated on the checkpoint fine-tuned from the epoch,
%		which is still under-fitting during pre-training, 
%		while that of the ESM-1b model is evaluated on the checkpoint fine-tuned from about 50\textit{th} epoch.
		$^*$The  PMLM-xl checkpoint fine-tuned from is pre-trained by a half of GPU time as that of the ESM-1b model does.}
	\label{tab:esm-lrange}
	\vspace{0.5em}
	\begin{tabular}{c|c|c|c|c}
		\hline
		\textsc{Model} & \textsc{\#Params} & \textsc{Data} & \textsc{Pre-train Task} & \textsc{Result}\\ 
		\hline\hline
		TAPE Transformer & 38M & Pfam & MLM & 23.2 \\ 
		UniRep & 18M & UR50 & Autoregressive & 21.9 \\
		SeqVec & 93M & UR50 & Autoregressive & 29.0 \\
		\hline
		ESM Transformer-12 & 85M & UR50 & MLM & 37.7 \\ 
		PMLM-base (ours) & 87M & UR50  & MLM + PMLM & 41.6 \\ % 41.6 by Liang, 46.1 by Fusong
		\hline
		ESM Transformer-34 & 669M & UR50  & MLM & 50.2 \\ 
		ESM-1b & 650M & UR50  & MLM & 56.9 \\ 
%		\hline
%		MLM-base (+ Linear) & 85M & Pfam & MLM & 19.8 \\ 
%		PMLM-base (+ Linear) & 85M & Pfam & MLM + PMLM & 25.2 \\ 
%		MLM-base (+ Linear) & 85M & UR50 & MLM & 36.9 \\ 
%		PMLM-base (+ Linear) (90) & 85M & UR50  & MLM + PMLM & 39.3 \\
%		PMLM-base$^+$ (+ Linear)  (90)  & 87M & UR50  & MLM + PMLM &  39.2 \\ 
%		PMLM-base$^+$ (+ ResNet-8)  (90) & 87M & UR50  & MLM + PMLM & 41.6 \\ 
%		PMLM-base (+ Linear) & 85M & UR50  & MLM + PMLM & 39.3 \\
%		PMLM-base$^+$ (+ Linear)  & 87M & UR50  & MLM + PMLM &  39.2 \\ 
%		PMLM-base$^+$ (+ ResNet-8) & 87M & UR50  & MLM + PMLM & \textbf{41.6} \\ 
%		\hline
%		PMLM-large (+ Linear)  & 252M & UR50  & MLM + PMLM & 51.3 \\ 
%		PMLM-large$^+$ (+ Linear) & 252M & UR50  & MLM + PMLM & 52.3 \\ 
%		PMLM-large$^+$ (+ ResNet-8) & 252M & UR50  & MLM + PMLM & 54.1 \\ 
		PMLM-xl (ours) & 715M & UR50  & MLM + PMLM & \, \textbf{59.9}$^*$ \\ 
		\hline
\end{tabular}
\end{table}

%\begin{table}[h!]
%	\centering
%	\TableSize{}
%	\caption{Medium-/Long- Range Precision@$L/5$ on RaptorX Contact Prediction Test Set}
%	\label{tab:esm-mlrange}
%	\vspace{0.5em}
%	\begin{tabular}{c|c|c|c|c}
%		\hline
%		\textsc{Model} & \textsc{\#Params} & \textsc{Data} & \textsc{Pre-train Task} & \textsc{Result}\\ 
%		\hline\hline
%		PMLM-base (+ Linear) & 85M & Pfam & MLM + PMLM & 45.1 \\ 
%		PMLM-base (+ Linear) & 85M & UR50  & MLM + PMLM & 64.9 \\ 
%%		PMLM-large (39) & 252M & UR50  & MLM + PMLM & 78.2 \\ 
%		PMLM-large (+ Linear) & 252M & UR50  & MLM + PMLM & 78.2 \\ 
%		\hline
%	\end{tabular}
%\end{table}

\subsection{Ablation Study}

To study the factors which influence the model performance, a systemic ablation study is conducted.

\subparagraph*{Pre-training Data}

Pfam is a dataset with protein sequences from a few thousand families, 
while UR50 consists of much more diverse sequences,
i.e., sequences from UniRef with 50\% sequence identity.
In other words, UR50 is more diverse than Pfam and representative for more sequences.
As shown in Table~\ref{tab:tape-mlrange},
when PMLM-base is pre-trained on UR50, which consists of sequences with higher diversity, instead of Pfam,
the performance gain for the task is more than $10\%$, increased from $47.2$ to $57.7$.
As shown in Table~\ref{tab:esm-lrange-ablation},
%The precision score of PMLM-base (+ Linear) pre-trained on UR50 is $14\%$ higher than that pre-trained on Pfam,
%further validate the impact of the dataset.
%A similar gap can be observed from Table~\ref{tab:esm-mlrange},
%by comparing the PMLM-base encoders pre-trained on Pfam and UR50.

\begin{table}[h!]
	\centering
	\TableSize{}
	\caption{Ablation Study on RaptorX Contact Prediction Test Set (Long-Range Precision@$L$)}
	\label{tab:esm-lrange-ablation}
	\vspace{0.5em}
	\begin{tabular}{c|c|c|c|c}
		\hline
		\textsc{Model} & \textsc{\#Params} & \textsc{Data} & \textsc{Pre-train Task} & \textsc{Result}\\ 
		\hline\hline
		MLM-base (+ Linear) & 85M & Pfam & MLM & 19.8 \\ 
		PMLM-base (+ Linear) & 85M & Pfam & MLM + PMLM & 25.2 \\ 
		\hline
		MLM-base (+ Linear) & 85M & UR50 & MLM & 36.9 \\ 
		%		PMLM-base (+ Linear) (90) & 85M & UR50  & MLM + PMLM & 39.3 \\
		%		PMLM-base$^+$ (+ Linear)  (90)  & 87M & UR50  & MLM + PMLM &  39.2 \\ 
		%		PMLM-base$^+$ (+ ResNet-8)  (90) & 87M & UR50  & MLM + PMLM & 41.6 \\ 
		% PMLM-base (+ Linear) & 85M & UR50  & MLM + PMLM & 39.3 \\
		% PMLM-base$^+$ (+ Linear)  & 87M & UR50  & MLM + PMLM &  39.2 \\ 
		% PMLM-base$^+$ (+ ResNet-8) & 87M & UR50  & MLM + PMLM & \textbf{41.6} \\ 
		PMLM-base (+ Linear)  & 87M & UR50  & MLM + PMLM &  39.2 \\ 
		PMLM-base (+ ResNet) & 88M & UR50  & MLM + PMLM & 41.6 \\ 
		%		\hline
		%		PMLM-large (+ Linear)  & 252M & UR50  & MLM + PMLM & 51.3 \\ 
		%		PMLM-large$^+$ (+ Linear) & 252M & UR50  & MLM + PMLM & 52.3 \\ 
		% 		PMLM-large$^+$ (+ ResNet-8) & 252M & UR50  & MLM + PMLM & 54.1 \\ 
		PMLM-large (+ ResNet) & 252M & UR50  & MLM + PMLM & 54.1 \\ 
		PMLM-xl (+ ResNet) & 715M & UR50  & MLM + PMLM &  59.9 \\ 
		\hline
	\end{tabular}
\end{table}

\subparagraph*{Model Size}

Model under-fitting is observed in both ESM model pre-training and ours.
A straightforward way to tackle this is by increasing the model size.
To examine this factor, we compare two models with different sizes pre-trained on the same data,
i.e. PMLM-base and PMLM-large.
As shown in Table~\ref{tab:tape-mlrange}, when PMLM-base and PMLM-large are both pre-trained on UR50,
the performance gap is about $14\%$ between $57.7$ (PMLM-base) and $71.7$ (PMLM-xl),
illustrating that increasing model size indeed benefits the performance for contact prediction.
Performance increases by more than $18\%$ as shown in Table~\ref{tab:esm-lrange-ablation}
when comparing PMLM-xl (+ ResNet) with PMLM-base (+ ResNet). Both are pre-trained on UR50.

%\subparagraph*{Pair Representation}
%
%Since different representations from the encoder may be used for fine-tuning, 
%we also conducted a group of experiments to study the difference.
%As shown in Table~\ref{tab:tape-mlrange}, for the pre-trained PMLM-large on UR50,
%when the fine-tuning process adopts $\mathcal{M}^+$ (the output of the first layer in the pair prediction head) 
%instead of $\mathcal{M}$ (the output of the encoder), the performance is increased from $64.3$ to $65.0$.
%But it does not always improve the performance, 
%e.g., PMLM-base$^+$ for % TAPE-Contact shown in Table~\ref{tab:tape-mlrange} and
%PMLM-base$^+$ (+ Linear) for RaptorX-Contact shown in Table~\ref{tab:esm-lrange},
%the final performance is close to that of PMLM-base.
%However, since no obvious performance drop is witnessed, 
%it is still a good choice to fine-tune on $\mathcal{M}^+$ for downstream tasks.

\subparagraph*{Downstream Task Module}

On top of the pre-trained encoder, there are various methods to fine-tune the model for downstream tasks.
To study the impact of this factor, we also compare the pre-trained encoder PMLM-base with two different
downstream modules, namely, a simple linear layer (+ Linear) and an eight-layer ResNet with convolution blocks (+ ResNet).
As shown in Table~\ref{tab:esm-lrange},
by comparing the performance of PMLM-base (+ Linear) and PMLM-base (+ ResNet),
we can see that the precision score is increased from $39.2$ to $41.6$,
indicating that a more expressive model can be trained to improve the performance on the ever pre-trained encoder.

\subparagraph*{Other Downstream Tasks}

\begin{table}[h!]
	\centering
	\TableSize{}
	\caption{Secondary Structure Prediction on the TAPE Benchmark}
	\label{tab:ssp-tape}
	\vspace{0.5em}
	\begin{tabular}{c|c|c|c}
		\hline
		\textsc{Model} & \textsc{CB513} & \textsc{CASP12} & \textsc{TS115} \\ 
		\hline\hline
		TAPE Transformer & 0.730 & 0.710 & 0.770  \\ 
		PMLM-base (pre-trained on Pfam, ours) & 0.728 & 0.706 & 0.771  \\ 
		\hline
	\end{tabular}
\end{table}

\begin{table}[h!]
	\centering
	\TableSize{}
	\caption{Remote Homology Prediction on the TAPE Benchmark}
	\label{tab:rh-tape}
	\vspace{0.5em}
	\begin{tabular}{c|c|c|c}
		\hline
		\textsc{Model} & \textsc{Fold} & \textsc{Superfamily} & \textsc{Family} \\ 
		\hline\hline
		TAPE Transformer & \textbf{0.210} & 0.340 & 0.880  \\ 
		PMLM-base (pre-trained on Pfam, ours) & 0.199 & \textbf{0.446} & \textbf{0.946}  \\ 
		\hline
	\end{tabular}
\end{table}

To evaluate our pre-trained encoder on other tasks, we also compared the performance of PMLM-base and the TAPE Transformer for 
Secondary Structure Prediction and Remote Homology Prediction on the TAPE benchmark,
the results in terms of accuracy scores are listed in Tables~\ref{tab:ssp-tape} and \ref{tab:rh-tape}.
The performance numbers of PMLM-base for secondary structure prediction are quite close to that of the TAPE baseline Transformer, 
illustrating that the proposed model does not hurt the performance with the additional loss.
For Remote Homology Prediction, the performance gap varies on the fold-level, superfamily-level, and family-level holdout test sets.
The performance of PMLM-base is slightly worse than that of the TAPE Transformer on the fold-level test set,
however, much better on the superfamily-level and family-level test sets up to 10\%,
showing the potential improvement from the extracted co-evolutionary information on other tasks.

%\section{Conclusion and Discussion}
\section{Discussion}
\label{conclusion}

In this paper, we propose a novel model named pairwise masked language model 
for the protein sequence encoder to capture co-evolutionary information during pre-training.
The pre-trained model generates a better representation for global structure by applying this model.
Our result shows that the proposed method significantly improves
the performance of contact prediction compared to the baselines.
Meanwhile, the proposed model surpasses the MSA baseline on the TAPE contact benchmark.
%
%\subsection{Discussion}
%
Although the performance is encouraging, the potential capability of the proposed model, i.e. PMLM, is not fully developed, 
neither it is the only way to extract the co-evolutionary information from the sequences.
For example, an observation is that increasing the mask probability for the tokens
might improve the representation for PMLM.

Although it sheds light on the single sequence based methods for representation learning for proteins,
more endeavors are needed to push forward the pre-training methods for protein sequences.
Following the idea of PMLM, supervision from multiple residues, 
e.g. a Triple Masked Language Model, might also be helpful for pre-training the representation. 
%Inspired by AlphaFold2,  the embeddings for two residues are correlated to the third one within the same sequence.
The metaphor is that three points in Euclidean space follows the triangle in-equations inspired by AlphaFold2.
%Although the learned representation space might not be restrictively Euclidean,
%it still indicates that residue triples in the sequence 
%is a reasonable option for pre-training the encoder.
However, the storage and computation cost will be cubic to the count of the masked tokens, 
thus how to efficiently pre-train the model might be an important issue to tackle,
which will be left to future work.

\section{Acknowledgments}

We would like to thank Dahai Cheng for his valuable work on initial analysis.

%\newpage

\bibliography{references}
\bibliographystyle{iclr2022_conference}

%\appendix
%\input{appendix}
%You may include other additional sections here.

\end{document}